%% file: ijcai25.tex
\documentclass{article}
\usepackage{ijcai25}
    
\usepackage{times}
\usepackage{soul}
\usepackage{url}
\usepackage[hidelinks]{hyperref}
\usepackage[utf8]{inputenc}
\usepackage[small]{caption}
\usepackage{graphicx}
\usepackage{amsmath}
\usepackage{amsthm}
\usepackage{booktabs}
\usepackage{algorithm}
\usepackage{algorithmic}
\usepackage[switch]{lineno}
\usepackage{xcolor}
\usepackage{colortbl}
\usepackage{makecell}
\usepackage[normalem]{ulem}
\useunder{\uline}{\ul}{}
\usepackage[symbol]{footmisc}

\title{Scan-and-Print: Patch-level Data Summarization and Augmentation \\ for Content-aware Layout Generation in Poster Design}

\author{
    HsiaoYuan Hsu and Yuxin Peng\thanks{Corresponding author.}
    \affiliations
    Wangxuan Institute of Computer Technology, Peking University
    \emails
    kslh99@stu.pku.edu.cn, pengyuxin@pku.edu.cn
}

\begin{document}

\maketitle
\input{sec/0_abstract}
\input{sec/1_intro}
\input{sec/2_related}
\input{sec/3_method}
\input{sec/4_experiments}
\input{sec/5_conclusion}
\input{sec/X_appendix}
\input{supp_sec/0_intro}
\input{supp_sec/1_results}
\input{supp_sec/2_samples}

\section*{Acknowledgments}
This work was supported by the grants from the National Natural Science Foundation of China (62525201, 62132001, 62432001) and Beijing Natural Science Foundation (L247006).

\bibliographystyle{named}
\bibliography{ijcai25}

\clearpage

\end{document}

%% file: sec/0_abstract.tex
\begin{abstract}
In AI-empowered poster design, content-aware layout generation is crucial for the on-image arrangement of visual-textual elements, \textit{e.g.}, logo, text, and underlay.
To perceive the background images, existing work demanded a high parameter count that far exceeds the size of available training data, which has impeded the model's \textit{real-time performance} and \textit{generalization ability}.
To address these challenges, we proposed a patch-level data summarization and augmentation approach, vividly named \textbf{Scan-and-Print}.
Specifically, the scan procedure selects only the patches suitable for placing element vertices to perform fine-grained perception efficiently.
Then, the print procedure mixes up the patches and vertices across two image-layout pairs to synthesize over 100\% new samples in each epoch while preserving their plausibility.
Besides, to facilitate the vertex-level operations, a vertex-based layout representation is introduced.
Extensive experimental results on widely used benchmarks demonstrated that Scan-and-Print can generate visually appealing layouts with \textit{state-of-the-art} quality while dramatically reducing computational bottleneck by 95.2\%.
The project page is at \href{https://thekinsley.github.io/Scan-and-Print/}{https://thekinsley.github.io/Scan-and-Print/}.
\end{abstract}

%% file: sec/1_intro.tex
\section{Introduction}

\begin{figure}
    \centering
    \includegraphics[width=1\linewidth]{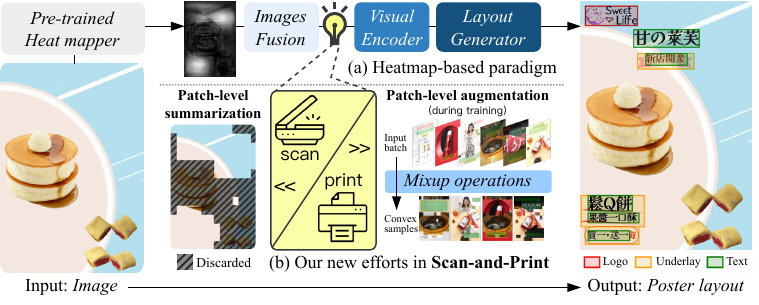}
    \vspace{-1.9em}
    \caption{Content-aware layout generation task. (a) Heatmap-based paradigm. (b) Our new efforts: data summarization for efficient image perception and data augmentation for enhanced model generalization, aiming for real-time, robust performance.}
    \label{fig:topic}
    \vspace{-1.3em}
\end{figure}

Integrating artificial intelligence (AI) with art and creativity has emerged as a pivotal trend in the design domain.
According to a survey of a mainstream design platform with over 185 million monthly users, 90\% of respondents agreed that AI has improved their work \cite{canva_ai}.
Among these advancements, content-aware layout generation plays a crucial role in automating poster design \cite{lin-2023-ACMMM-autoposter,yang-2023-ACMMM-videothumb,weng-2024-CVPR-desigen,wang-2024-ACMMM-prompt2poster} by indicating the arrangement of visual-textual elements, \textit{e.g.}, logo, text, and underlay, on the background image, as shown in Fig. \ref{fig:topic}.

Despite the increasing attention to this valuable task \cite{Min-2022-IJCAI-CGL,hsu-2023-CVPR-posterlayout,xu-2023-CVPR-PDA,horita-2024-CVPR-RALF}, existing methods faced a high computational bottleneck in perceiving images.
Taking the current \textit{state-of-the-art} (SOTA) method, RALF \cite{horita-2024-CVPR-RALF}, as an example, even when neglecting the time required for image-retrieval augmentation \cite{fu-2024-NIPS-dreamsim}, it took an average of 385 ms per single inference on an NVIDIA A40 GPU.
The saliency detection in preprocessing accounted for 55 ms, while more critically, the image encoders during generation occupied over 69.5\% of the model parameters.
This not only poses significant challenges for \textit{real-time performance}, but the scarcity of training data relative to the model's capacity also impairs the model's \textit{generalization ability}.

To relieve the inefficiency and overfitting issues in the field, we proposed \textbf{Scan-and-Print}, an auto-regressive model accompanied by patch-level data summarization and augmentation.
Specifically, (a) the scan procedure selects only the few patches from an input image that are predicted to have a high probability of placing element vertices, thereby concentrating computational resources on the most applicable areas.
Then, (b) the print procedure synthesizes augmented samples by mixing the patches and vertices from two image-layout pairs, effectively increasing both the size and diversity of datasets.
We also introduced a new (c) vertex-based layout representation to facilitate vertex-level mixup operations.

We conducted extensive experiments on the public benchmarks \cite{Min-2022-IJCAI-CGL,hsu-2023-CVPR-posterlayout} and demonstrated that Scan-and-Print has achieved new SOTA performance.
Compared to RALF \cite{horita-2024-CVPR-RALF}, it has drastically reduced 95.2\% of the FLOPs in image encoders.
The synthesized data have consistently shown positive impacts even when the augmentation rate reached more than 100\%, highlighting their plausibility and usability.
Besides, we demonstrated the adaptability of Scan-and-Print to user-specified constraints, particularly beneficial for the real-world poster design workflow.

The contribution of this work are summarized as follows:
\begin{itemize}
    \item A data summarization approach (Scan) selects only the few patches suitable for arranging vertices of layout elements to efficiently perceive input image content.
    \item A data augmentation approach (Print) mixes the patches and vertices from two image-layout pairs to synthesize extensive new plausible samples at a low cost.
    \item A vertex-based layout representation (VLR) models fine-grained geometric properties to facilitate delicate vertex-level mixup operations across layouts.
    \item Comprehensive evaluations verify the practical application value of Scan-and-Print, being the first in the field to focus on reducing computational complexity.
\end{itemize}

%% file: sec/2_related.tex
\section{Related Work}
\subsection{Content-aware Layout Generation}
Different from general layout tasks \cite{Li-2020-TPAMI-LayoutGAN,weng-2023-IJCAI-learnsample}, content-aware layout generation additionally takes into account the given background image, thus possessing high application value in the field of AI-empowered design \cite{wang-2024-ACMMM-prompt2poster,weng-2024-CVPR-desigen}.
%
%
Pioneered in CGL-GAN \cite{Min-2022-IJCAI-CGL} and PKU PosterLayout \cite{hsu-2023-CVPR-posterlayout} to establish the heatmap-based paradigm, \textit{i.e.}, utilizing object saliency \cite{Li-2021-TMM-BASNet,qin-2022-ECCV-ISNet} or spatial density \cite{hsu-2023-icig-densitylayout} maps to enhance the awareness of image composition.
Although CGL and PKU contributed valuable datasets, the total size of training data remains scarce around 70.5k samples, urging for \textit{heuristic techniques} or \textit{data augmentation approaches} to improve model performance.

Inspired by the prior design experiences \cite{Guo-2021-CHI-Vinci,Li-2020-TVCG-attrlayoutgan}, DS-GAN \cite{hsu-2023-CVPR-posterlayout} organized layout elements in a motivation-aligned order to mine patterns in the data more effectively during GANs' training.
\cite{chai-2023-ACMMM-twostage} employed a general diffusion model \cite{chai-2023-CVPR-layoutdm} for content-aware tasks by incorporating predefined aesthetic constraints and a saliency-aware layout plausibility ranker.
In another way, RALF \cite{horita-2024-CVPR-RALF} incorporated retrieval augmentation by searching for the nearest neighbors of the input image \cite{fu-2024-NIPS-dreamsim} and using their layout features as additional input for autoregressive models.
LayoutPrompter \cite{lin-2023-NIPS-layoutprompter} coarsely extracted the minimum bounding rectangle from the saliency maps and retrieved layout examples to enable the in-context learning of LLMs.
PosterLlama \cite{seol-2024-ECCV-posterllama} synthesized new image-layout samples by a depth-guided image generation with refined text descriptions \cite{zhang-2023-ICCV-ControlNet,li-2023-ICML-blip2}, while maintaining the corresponding layouts unchanged, to fine-tune DINOv2 \cite{oquab-2023-arXiv-dinov2,zhu-2023-arxiv-minigpt} and CodeLlama-7B \cite{roziere-2023-arXiv-codellama,hu-2021-arXiv-lora} successively.

However, along with these efforts, the size of model parameters, especially image encoders, often exceeds that of the available training data.
This leads to significant challenges in real-time performance and generalization ability.
In light of this, we devote this work to a compact model with selective \textit{scan} and efficient data augmentation--  \textit{print}.

\subsection{Patch-level Data Augmentation}

Data augmentation is a crucial regularization method to enhance model generalization by artificially increasing the size and diversity of training data.
In image understanding, traditional operations such as random cropping and flipping have been widely used.
An advanced field is mixing-based data augmentation, which synthesizes new data by combining multiple samples.
\textit{E.g.}, Mixup \cite{zhang-2018-ICLR-mixup} randomly drew two image-label pairs $(x_{i}, y_{i}), (x_{j}, y_{j})$ and linearly interpolated them to obtain the convex combination $(\tilde{x},\tilde{y})$.

Upon the success of Mixup, patch-level mixing approaches have been developed to create more realistic samples that preserve the spatial structure of the images.
CutMix \cite{yun-2019-ICCV-cutmix} cropped a rectangular region of $x_i$ and filled the corresponding part of $x_j$, showing particular advantages in localization tasks.
As random selection sometimes results in mixed patches lacking supervisory information, saliency detection has been introduced into the process.
Puzzle Mix \cite{kim-2020-ICML-puzzlemix} formulated an optimization problem, maximizing the exposed saliency, to jointly determine the size of the mixing mask and its spatial offset between the samples.
SaliencyMix \cite{uddin-2020-ICLR-saliencymix} straightforwardly selected the peak salient area within $x_j$ and pasted it on $x_i$.
In contrast, Co-Mixup \cite{kim-2021-ICLR-comixup} sophisticatedly paired the samples from the mini-batch to obtain the largest possible accumulated saliency regions.
GuidedMix \cite{kang-2023-AAAI-guidedmixup} further sped up this complicated process by splitting apart the pairing and mask determination.

These efforts have consistently enhanced model robustness and generalization ability in deep learning-based computer vision methods \cite{bochkovskiy-2020-arXiv-yolov4,bang-2022-IJCAI-logitmix}.
However, the labels $y$ in previous tasks are often simplistic, \textit{e.g.}, one-hot vectors, which fail to serve the complex, hierarchical structure of layout elements.
In light of this, we discussed the vertex-level label mixup to mitigate this gap.

%% file: sec/3_method.tex
\begin{figure*}[t]
    \centering
    \includegraphics[width=0.99\linewidth]{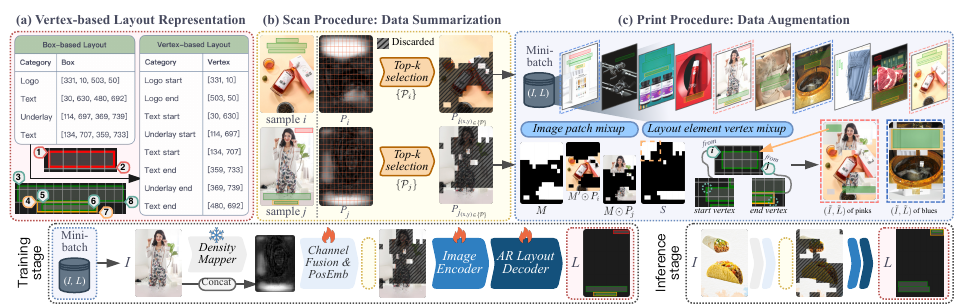}
    \vspace{-0.5em}
    \caption{An overview of Scan-and-Print. Preliminarily, (a) represents layout $L$ based on the precise geometric properties, \textit{i.e.}, vertices, and grouping relationship, \textit{i.e.}, underlays, to facilitate the following fine-grained procedures; (b) efficiently `scans' the input image $I$ to perceive only the few patches suitable for arranging element vertices; (c) `prints' augmented samples $(\Tilde{I}, \Tilde{L})$ as extra training data by mixing patches and vertices across different pairs within the mini-batch to enhance the generalization ability of the autoregressive model.}
    \label{fig:pipeline}
    \vspace{-1em}
\end{figure*}

\section{The Proposed Approach: Scan-and-Print}
Considering the computational overhead and generalization issues of current content-aware layout generation approaches, we propose Scan-and-Print.
It is an autoregressive model that achieves \textit{efficiency in parameter count, image perception, and data augmentation}.
An overview is shown in Fig. \ref{fig:pipeline}.
Briefly, we propose (a) vertex-based layout representation (\textbf{VLR}) to capture the fine-grained structure of elements and facilitate the two procedures, namely, (b) \textbf{scan} that identifies and perceives a small size $k$ of image patches applicable for element placement, and (c) \textbf{print} that synthesizes new training samples $(\Tilde{I}, \Tilde{L})$ by mixing patches and vertices across two image-layout pairs $(I_{i}, L_{i})$ and $(I_{j}, L_{j})$ within the given mini-batch.

\subsection{Vertex-based Layout Representation}

\begin{algorithm}[t]
\caption{Vertex-based layout representation (VLR)}
    \label{alg:ver}
    \textbf{Input}: Layout $L=\{e_i\}_i^n=\{(c_i, (x_{\text{c}_i}, y_{\text{c}_i}, w_i, h_i))\}_i^n$ \\
    \textbf{Parameter}: Category ID of underlay $c_{und}$ \\
    \textbf{Output}: Start-end vertex tensor $V=\{(c_i, (x_i, y_i))\}_i^{2n}$ \\
    \vspace{-1.2em}
    \begin{algorithmic}[1] 
        \STATE $(x_l, y_t, x_r, y_b) \gets \textsc{cxcywh-to-xyxy}((x_{\text{c}}, y_{\text{c}}, w, h))$
        \STATE $G \gets$ \textsc{\textbf{group-element-id}}$(c_{und}, c, (x_l, y_t, x_r, y_b))$
        \STATE $C \gets \textsc{repeat}(c, n \rightarrow (n\ 2))$
        \FOR{i = 1 \TO n}
        \STATE $C_{2i} \gets 2C_{2i}$
        \STATE $C_{2i+1} \gets C_{2i} + 1$ \ \ $\triangleright$ Increment for end vertex
        \ENDFOR
        \STATE $X \gets \textsc{rearrange}((x_l, x_r), n\ (2\ d) \rightarrow (n\ 2)\ d)$
        \STATE $Y \gets \textsc{rearrange}((y_t, y_b), n\ (2\ d) \rightarrow (n\ 2)\ d)$
        \STATE $W \gets 0.01 \times X + Y $ \ \ $\triangleright$ Weights for top-left ordering
        \STATE $A \gets$ \textsc{\textbf{arrange-severtex-id}}$(G, W)$
        \STATE $V \gets \{(c_{A[i]}, (x_{A[i]}, y_{A[i]}))\}_i^{2n}$
        \STATE \textbf{return} $V$
    \end{algorithmic}
    \label{alg:vlr}
\end{algorithm}

Conventionally, a layout $L$ is represented as a set of bounding boxes, each described by its center coordinates and size, \textit{i.e.}, $\{e_i\}_i^n\!=\!\{(c_i, (x_{\text{c}_i}, y_{\text{c}_i}, w_i, h_i))\}_i^n$.
With its simplicity comes the difficulty of modeling fine-grained spatial relationships and manipulating geometric properties precisely, leading to poor graphic quality, such as misalignment \cite{Min-2022-IJCAI-CGL}.
To this end, we propose a more direct representation based on the vertices of boxes.
As depicted in Algorithm \ref{alg:vlr}, VLR first derives the top-left and bottom-right coordinates $(x_{\text{l}}, y_{\text{t}}, x_{\text{r}}, y_{\text{b}})$ of elements, and then performs \textsc{\textbf{group-element-id}} to construct ID trees, which explicitly reflects the hierarchical structure in $L$.
Subsequently, the categories and coordinates are transformed into the attributes of start-end vertices.
Inspired by \cite{hsu-2023-CVPR-posterlayout}, \textsc{\textbf{arrange-severtex-id}} is performed to sort and obtain the vertex tensor $V$ but only considers the reading order and grouping relationship.
%
For a smooth reading experience, the definitions of the two functions are postponed to Appendix \ref{app:function}.

\subsection{Scan Procedure: Data Summarization}
\paragraph{Image encoder with patch selection.}
In the target context, image understanding essentially boils down to a binary determination of whether specific areas are suitable for placing layout elements.
Relative to its limited complexity, existing methods are prone to employing disproportionately heavy image encoders, occupying up to 82.4\% of the model's trainable parameters \cite{hsu-2023-CVPR-posterlayout}.
This insight naturally leads us to a compact encoder focusing on a small proportion within the input image $I$.
Preliminarily, a parameter-efficient density mapping network is pre-trained semi-supervised as in \cite{hsu-2023-icig-densitylayout} to identify top-$k$ patches $\{P_{(\text{x},\text{y})}\! \mid\! (\text{x},\text{y})\! \in\! \{\mathcal{P}\}\}$ with the highest scores for placing element vertices, where $\vert \{\mathcal{P}\} \vert\! =\! k\! \ll\!$ the number of patches $p^2$.
This selection is performed after the positional embedding, so the subsequent encoder retains the original spatial information of $P_{(\text{x},\text{y})}$ as well, which is crucial for understanding global relationships.

\paragraph{Image-to-layout alignment and autoregressive decoder.}
Following \cite{horita-2024-CVPR-RALF}, a layout tokenizer is utilized to quantize $x$ and $y$ into 128 bins.
To bridge the modality gap \cite{chen-2024-IJCAI-VTPH} between visual and geometric features, a two-layer FFN is inserted after the last layer of the encoder.
Finally, an autoregressive decoder, trained through the next token prediction objective, sequentially predicts $6n$ tokens-- detokenized as $n$ elements afterward.

\subsection{Print Procedure: Data Augmentation}
Based on the concept of Mixup \cite{zhang-2018-ICLR-mixup}, image-layout pairs $(I_{i}, L_{i})$ and $(I_{j}, L_{j})$ are picked from the given mini-batch to construct a convex combination $(\Tilde{I}, \Tilde{L})$, considering their patch indices $\{\mathcal{P}_i\}$ and $\{\mathcal{P}_j\}$.
In total, $\alpha$ new samples are created per mini-batch during training, effectively improving the model's generalization ability.

\begin{table*}[t]
\centering

\begin{minipage}{0.76\textwidth}
\resizebox{0.98\textwidth}{!}{%
\begin{tabular}{l|c|cccc|ccc}
\toprule
Method         & \#Params & $Ove \downarrow$ & $Ali \downarrow$ & $Und_l \uparrow$ & $Und_s \uparrow$ & $Uti \uparrow$  & $Occ \downarrow$ & $Rea \downarrow$ \\ 
\midrule
\textit{\small LLM-based} & & & & & & & \\
LayoutPrompter        & 8B  & 0.0010          & 0.0026          & 0.4054          & 0.1621          & 0.2025          & 0.2539          & 0.0412          \\
PosterLLaMa$^\ddag$   & 7B  & 0.0006          & 0.0006          & 0.9986          & 0.9917          & 0.1764          & 0.1630          & 0.0285          \\
\midrule
\textit{\small Non-LLM-based} & & & & & & & \\
CGL-GAN               & 41M & 0.1010          & 0.0048          & 0.7326          & 0.2743          & 0.1693          & 0.2105          & 0.0327          \\
DS-GAN                & {\ul 30M} & 0.0248          & 0.0046          & 0.7859          & 0.4676          & {\ul 0.1832}    & 0.1894          & 0.0320          \\
ICVT                  & 50M & 0.2786          & 0.0480          & 0.4939          & 0.3549          & 0.1050          & 0.2686          & 0.0347          \\
LayoutDM$^\dag$          & 43M & 0.1638          & {\ul 0.0029} & 0.5987          & 0.3695          & 0.1475          & 0.1504          & 0.0264          \\
AutoReg               & 41M & 0.0218          & 0.0052          & 0.7053          & 0.3537          & 0.1449          & 0.1535          & 0.0274          \\
RALF                  & 43M & {\ul 0.0175}    & 0.0069          & {\ul 0.9548}    & {\ul 0.8653}    & 0.1452          & {\ul 0.1328}    & {\ul 0.0231}    \\
Scan-and-Print (Ours) & \textbf{26M} & \textbf{0.0090}  & \textbf{0.0024}  & \textbf{0.9831}  & \textbf{0.9709}  & \textbf{0.1985} & \textbf{0.1162}  & \textbf{0.0181} \\
\bottomrule
\end{tabular}
}
\caption{Quantitative results on PKU PosterLayout dataset, \textit{unannotated} test split.}
\label{tab:res_pku_test}
\end{minipage}
\begin{minipage}{0.23\textwidth}
\vspace{-0.2em}
\includegraphics[width=1\textwidth]{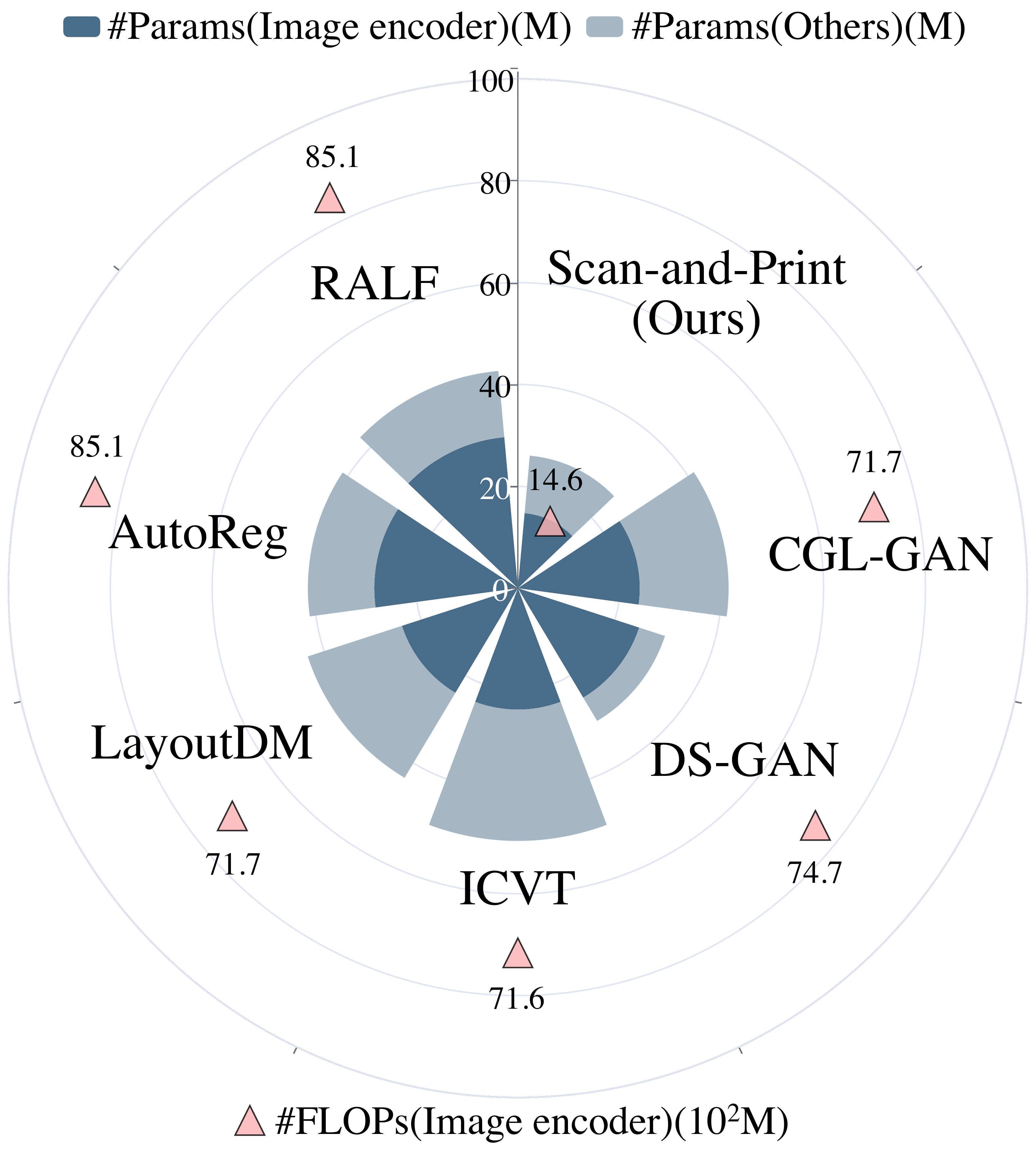}
\vspace{-1.8em}
\captionof{figure}{Comparisons of computational cost.}
\label{fig:cost}
\end{minipage}
\vspace{-1em}
\end{table*}

\paragraph{Image patch mixing.}
\label{subsec:ipm}
$I_i$ and $I_j$ are combined using a binary mask $M \in \{0, 1\}^{p \times p}$ and its complement $M'$, defined as:
\begin{subequations}
    \begin{align}
        M_{(\text{x},\text{y})} &= \begin{cases}
            1, & \text{if } (\text{x},\text{y}) \in \{\mathcal{P}_i\},  \\
            0, &  \text{otherwise},
            \end{cases}\\
        \Tilde{I} &= M' \odot I_i + M \odot I_j, \\
                  &= M' \odot \{P_{i(\text{x}\leq p,\text{y}\leq p)}\} + M \odot \{P_{j(\text{x}\leq p,\text{y}\leq p)}\},
    \end{align}
\end{subequations}
where $\odot$ denotes patch-wise multiplication.
This removes the applicable patches of $I_i$ and pastes on the corresponding patches of $I_j$, making $\Tilde{I}$ the challenging case containing fewer patches suitable for placing elements than the sources.

\paragraph{Element vertex mixing.}
A stricter mask $S^{p \times p}$ considering $\{\mathcal{P}_i \cap \mathcal{P}_j\}$ is initiated to constrain this process, ensuring that $\Tilde{L}$ is a plausible layout for the input $\Tilde{I}$, and then, all continuous regions $R$ with at least three available patches in $S$ is found by depth-first search.
Before mixing points in $(V_i, V_j)$, the longest common subsequence (LCS) of their categories $(C_i, C_j)$ is obtained as $\Tilde{C}$ of $\Tilde{L}$ by dynamic programming.
With indices $v_{i/j}\! = \!\{\textsc{index-of}(\text{LCS}_l)\textsc{in}
(V_{i/j})\}_l^{\textsc{length(\text{LCS})}}$
and regions $R$,
the points $(\Tilde{X}, \Tilde{Y})$ are abstracted as:
\begin{subequations}
    \begin{align}
        m &= \min(\textsc{length(\text{LCS})} / 2, \textsc{length(R)}), \\
        \Tilde{X} &= \begin{cases}
            \{X_{i}[l]\}_l^m, & \text{if } \textsc{is-start-category}(\Tilde{C}[l]),  \\
            \{X_{j}[l]\}_l^m, &  \text{otherwise},
            \end{cases} \\
        \Tilde{Y} &= \begin{cases}
            \{Y_{i}[l]\}_l^m, & \text{if } \textsc{is-start-category}(\Tilde{C}[l]),  \\
            \{Y_{j}[l]\}_l^m, & \text{otherwise}.
            \end{cases}
    \end{align}
\end{subequations}
This forms $\Tilde{V}$, where all starting points come from $V_i$, and all ending points come from $V_j$.
To increase randomness, $v_{i/j}$ and $R$ are shuffled carefully before this process.
Finally, the shifting operation is defined to move each pair $(V_s, V_e)$ of start-end points into the corresponding region $r$ while keeping their relative positions in the original patch, as:
\begin{subequations}
    \begin{align}
        V_s.x &= V_s.x\! \mod (I_{w} / p) + \textsc{left-top}(r).x, \\
        V_s.y &= V_s.y\! \mod (I_{h} / p) + \textsc{left-top}(r).y, \\
        V_e.x &= V_e.x\! \mod (I_{w} / p) + \textsc{right-bottom}(r).x, \\
        V_e.y &= V_e.y\! \mod (I_{h} / p) + \textsc{right-bottom}(r).y,
    \end{align}
\end{subequations}
where $(I_{w}, I_{h})$ is the input image size, and \textsc{left-top}, \textsc{right-bottom} return $(x_l,y_t)$ coordinates of the specified patch.

%% file: sec/4_experiments.tex
\section{Experiments}

\subsection{Datasets and Evaluation Metrics}
To evaluate the proposed Scan-and-Print, we conduct experiments on widely used e-commerce poster datasets, PKU PosterLayout \cite{hsu-2023-CVPR-posterlayout} and CGL \cite{Min-2022-IJCAI-CGL}.
Their train/annotated test/unannotated test splits are allocated following \cite{horita-2024-CVPR-RALF} to ensure a fair comparison with existing work.
Concretely, \textbf{PKU PosterLayout} contains 8,734/1,000/905 samples with three element types, which are logo, text, and underlay.
\textbf{CGL} contains 54,546/6,002/1,000 samples with four element types, where the extra one is embellishment.

\begin{table}[t]
    \centering
    \resizebox{0.99\linewidth}{!}{%
\begin{tabular}{@{\hspace{1pt}}l@{\hspace{3pt}}|cccc|ccc@{\hspace{1pt}}}
\toprule
Method                        & $Ove \downarrow$ & $Ali \downarrow$ & $Und_l \uparrow$ & $Und_s \uparrow$ & $Uti \uparrow$  & $Occ \downarrow$ & $Rea \downarrow$ \\
\midrule
LayoutPrompter                & 0.0026           & 0.0016           & 0.2693           & 0.1142           & 0.2008          & 0.4570           & 0.0644           \\
PosterLLaMa$^\ddag$           & 0.0014           & 0.0007           & 0.9971           & 0.9771           & 0.1032          & 0.4687           & 0.0555           \\
\midrule
CGL-GAN                       & 0.2668           & 0.0316           & 0.6774           & 0.1656           & 0.0554          & 0.4312           & 0.0512           \\
DS-GAN                        & 0.0991           & \textbf{0.0138}           & 0.7566           & 0.2810           & \textbf{0.1339}          & 0.4277           & 0.0526           \\
ICVT                          & 0.2045           & 0.1010           & 0.4357           & 0.2599           & 0.0360          & 0.4620           & {\ul 0.0397}     \\
LayoutDM$^\dag$                  & 0.0793           & 0.1822           & 0.6304           & 0.3853           & 0.0131          & 0.5438           & 0.0612           \\
AutoReg                       & 0.0577           & 0.0226           & 0.8848           & 0.7599           & {0.0572}          & 0.3839           & 0.0427           \\
RALF                          & {\ul 0.0273}     & {\ul 0.0189}     & {\ul 0.9756}     & {\ul 0.9315}     & {\ul 0.0601} & {\ul 0.3359}     & {\ul 0.0397}     \\
S-and-P (Ours)         & \textbf{0.0157}  & {0.0197}  & \textbf{0.9853}  & \textbf{0.9736}  & 0.0571          & \textbf{0.3356}  & \textbf{0.0323}  \\
\bottomrule
\end{tabular}
}
\vspace{-0.5em}
\caption{Quantitative results on CGL dataset, \textit{unannotated} test split.}
\vspace{-1em}
\label{tab:res_cgl_test}
\end{table}

\begin{figure*}[t]
    \centering
    \includegraphics[width=0.995\linewidth]{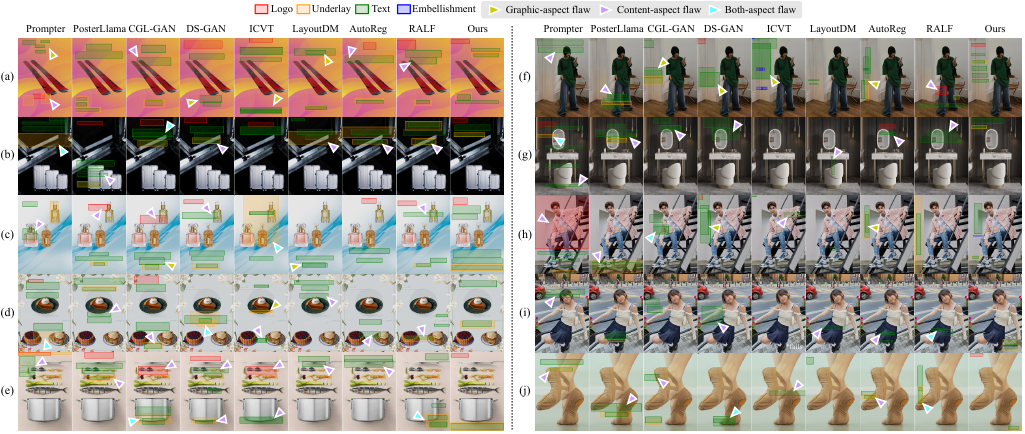}
    \vspace{-0.5em}
    \caption{Comparisons of visualized results on (a)-(e) PKU PosterLayout and (f)-(j) CGL datasets' \textit{unannotated} test splits.}
    \label{fig:vis}
    \vspace{-1.2em}
\end{figure*}

Following the above work, we evaluate layouts in two aspects, including (a) \textbf{graphic metrics}: overlay $Ove\!\!\downarrow$, alignment $Ali\!\!\downarrow$, loose, and strict underlay effectiveness $Und_l\!\!\uparrow$, $Und_s\!\!\uparrow$, and
(b) \textbf{content metrics}: space utilization $Uti\!\uparrow$, salient object occlusion $Occ\!\!\downarrow$, and readability $Rea\!\!\downarrow$.

\subsection{Implementation Details}
We implement the image encoder and layer decoder with the first 8 layers of ViT-S \cite{touvron-2022-ECCV-deit} and a 4-layer Transformer Decoder, respectively.
The size of the input image $I$ is (224, 224) and the embedding dimension is 384.
We set the batch size, epoch, learning rate of the encoder, and of the others as 128, 15, $5e^{-5}$, and $5e^{-4}$.
Considering the available data, the scan size $k$ is 96 and 48 for PKU PosterLayout and CGL, and the augmented sample size $\alpha$ is 256 and 16, creating approximately 17,644 and 6,816 samples per epoch.
All experiments are carried out on an NVIDIA A40 GPU.

\subsection{Comparison with State-of-the-arts}
We select approaches with open-sourced implementation as baselines, including 
the GAN-based CGL-GAN \cite{Min-2022-IJCAI-CGL}, DS-GAN \cite{hsu-2023-CVPR-posterlayout}, 
autoregression-based ICVT \cite{Cao-2022-ACMMM-ICVT}, AutoReg \cite{horita-2024-CVPR-RALF}, RALF \cite{horita-2024-CVPR-RALF}, 
diffusion model-based LayoutDM\footnote[2]{The extended version presented in \cite{horita-2024-CVPR-RALF}.} \cite{inoue-2023-CVPR-layoutdm}, and 
LLM-based PosterLlama\footnote[3]{With the released CodeLlama-7B weight tuned on \cite{hsu-2023-CVPR-posterlayout}, \cite{Min-2022-IJCAI-CGL}, and depth-guided augmented data.}\! \cite{seol-2024-ECCV-posterllama}, LayoutPrompter\footnote[4]{With the Llama3.1-8B weight \cite{dubey-2024-arXiv-llama3}.} \cite{lin-2023-NIPS-layoutprompter}.

\begin{table}[t]
    \centering
    \resizebox{0.99\linewidth}{!}{%
\begin{tabular}{@{\hspace{1pt}}l@{\hspace{3pt}}|cccc|cc@{\hspace{8pt}}c@{\hspace{1pt}}}
\toprule
Method                        & $Ove \downarrow$ & $Ali \downarrow$ & $Und_l \uparrow$ & $Und_s \uparrow$ & $Uti \uparrow$  & $Occ \downarrow$ & $Rea \downarrow$ \\
\midrule
\textcolor{gray!60}{Real data} & \textcolor{gray!60}{0.0010} & \textcolor{gray!60}{0.0038} & \textcolor{gray!60}{0.9955} & \textcolor{gray!60}{0.9896} & \textcolor{gray!60}{0.2238} & \textcolor{gray!60}{0.1193} & \textcolor{gray!60}{0.0109} \\
\midrule
LayoutPrompter                & 0.0017           & 0.0028           & 0.4085           & 0.1613           & 0.2104          & 0.2271           & 0.0309           \\
PosterLLaMa$^\ddag$           & 0.0008           & 0.0006           & 0.9999           & 0.9982           & 0.1812          & 0.1489           & 0.0177           \\
\midrule
CGL-GAN                       & 0.0966           & 0.0035           & 0.7854           & 0.3570           & 0.2065          & 0.1548           & 0.0191           \\
DS-GAN                        & 0.0261           & 0.0038           & 0.8350           & 0.5804           & 0.2078          & 0.1591           & 0.0199           \\
ICVT                          & 0.2572           & 0.0405           & 0.5384           & 0.3932           & 0.1161          & 0.2401           & 0.0259           \\
LayoutDM$^\dag$                  & 0.1562           & {\ul 0.0018}     & 0.6426           & 0.3873           & 0.1600          & 0.1432           & 0.0185           \\
AutoReg                       & 0.0187           & 0.0019           & 0.7863           & 0.4344           & 0.1994          & 0.1338           & 0.0164           \\
RALF                          & \textbf{0.0084}     & 0.0028           & \textbf{0.9808}  & {\ul 0.9201}     & {\ul 0.2137} & {\ul 0.1195}     & {\ul 0.0128{\small 4}}     \\
S-and-P (Ours)         & {\ul 0.0087}                & \textbf{0.0014}             & {\ul 0.9736}                & \textbf{0.9639}             & \textbf{0.2270}             & \textbf{0.1173}             & \textbf{0.0128{\small 1}}  \\
\bottomrule
\end{tabular}
}
\vspace{-0.5em}
\caption{Quantitative results on PKU PosterLayout dataset, \textit{annotated} test split.}
\vspace{-1.5em}
\label{tab:res_pku_valid}
\end{table}

\paragraph{Baseline comparison.}
As reported in Tab. \ref{tab:res_pku_test} and Tab. \ref{tab:res_cgl_test}, Scan-and-Print consistently achieves new SOTA performance across different benchmarks, especially on the severely data-scarce PKU PosterLayout.
Compared to existing LLM-based methods that represent layouts as structured language and utilize LLMs’ coding abilities, ours shows unprecedentedly comparable graphic effectiveness with only 26M parameters while significantly improving content metrics.
Particularly, it outperforms the SOTA approach, \textit{e.g.}, PosterLlama, by 28.7\% and 28.4\% in $Occ\!\downarrow$ across two benchmarks.
On the other hand, when compared to the non-LLM-based SOTA approach, \textit{e.g.}, RALF, it shows an overall superiority, especially improves $Und_s\!\uparrow$ by 12.2\% and 4.5\%.
These observations demonstrate that the proposed Scan-and-Print can generate visually appealing layouts, ensuring that (1) salient objects in the input image are not occluded and (2) complex structural elements are correctly organized.

Results on the annotated splits are also reported, as in 
Tab. \ref{tab:res_pku_valid} and Tab. \ref{tab:res_cgl_valid}.
We found the layouts generated by Scan-and-Print are already very close to the quality of ground truth data and even show better performance in $Ali\!\downarrow$ and $Occ\!\downarrow$.
While some metrics on CGL are slightly behind RALF, the overall performance of Scan-and-Print presents a good trade-off considering its reduced computational complexity, underlined in the next paragraph.


\begin{table}[t]
    \centering
    \resizebox{0.99\linewidth}{!}{%
\begin{tabular}{@{\hspace{1pt}}l@{\hspace{3pt}}|cccc|ccc@{\hspace{1pt}}}
\toprule
Method                        & $Ove \downarrow$ & $Ali \downarrow$ & $Und_l \uparrow$ & $Und_s \uparrow$ & $Uti \uparrow$  & $Occ \downarrow$ & $Rea \downarrow$ \\
\midrule
\textcolor{gray!60}{Real data} & \textcolor{gray!60}{0.0003} & \textcolor{gray!60}{0.0024} & \textcolor{gray!60}{0.9949} & \textcolor{gray!60}{0.9875} & \textcolor{gray!60}{0.1978} & \textcolor{gray!60}{0.1353} & \textcolor{gray!60}{0.0119} \\ \midrule
LayoutPrompter                & 0.0017           & 0.0030           & 0.3830           & 0.1740           & 0.1835                & 0.2380           & 0.0327           \\
PosterLLaMa$^\ddag$           & 0.0006           & 0.0006           & 0.9987           & 0.9890           & 0.1775                & 0.1647           & 0.0184           \\
\midrule
CGL-GAN                       & 0.2291           & 0.0123           & 0.6466           & 0.2281           & 0.1096          & 0.1811           & 0.0213           \\
DS-GAN                        & 0.0460           & {\ul 0.0022}     & 0.9081           & 0.6308           & 0.2408          & 0.1476           & 0.0181           \\
ICVT                          & 0.2453           & 0.0179           & 0.5150           & 0.3326           & 0.1488          & 0.1945           & 0.0211           \\
LayoutDM$^\dag$                  & 0.0184           & \textbf{0.0021}  & 0.9216           & 0.8159           & 0.1933          & 0.1369           & {\ul 0.0137}     \\
AutoReg                       & 0.0109           & 0.0023           & 0.9670           & 0.9171           & 0.1926          & {\ul 0.1250}     & 0.0190           \\
RALF                          & {\ul 0.0042}     & 0.0024           & \textbf{0.9912}  & \textbf{0.9756}  & \textbf{0.1969} & \textbf{0.1246}  & 0.0180           \\
S-and-P (Ours)         & \textbf{0.0034}             & 0.0023                      & {\ul 0.9701}                & {\ul 0.9639}                & {\ul 0.1957}                & 0.1336                      & \textbf{0.0126}  \\
\bottomrule
\end{tabular}
}
\vspace{-0.5em}
\caption{Quantitative results on CGL dataset, \textit{annotated} test split.}
\vspace{-1.3em}
\label{tab:res_cgl_valid}
\end{table}

\paragraph{Computational cost.}
The parameter counts and FLOPs of different approaches are reported in Fig.\! \ref{fig:cost}.
As observed, the total \#Params in Scan-and-Print is only 61\% of that in RALF, even less than \#Params of RALF's image encoders, which is $2\times$ the Scan-and-Print's image encoder.
This efficiency is further strengthened by the proposed \textit{scan} procedure.
Specifically, when the size $k$ is set to 96, our image encoder consumes only 1.46G FLOPs, which is 17\% of the 8.51G FLOPs required by RALF.
More exploration of $k$ values will be reported in the ablation study.

\paragraph{Visualized results.}
Fig. \ref{fig:vis} shows the layouts generated by different methods.
The results illustrate that Scan-and-Print
specializes in organizing combinations of elements rarely or never seen in datasets to suit applicable areas of diverse sizes and distributions, which is the charm of Print-- the mixing-based data augmentation, while Scan ensures these elements are properly placed and reduces undesirable occlusions.
Specifically, seeing the images in the first column, \textit{i.e.}, (a) and (f), although their only objects leave enough available spaces, layouts generated by existing methods still have minor flaws.
In contrast, ours generates nearly perfect ones that actively utilize most spaces, creating more and better-organized elements.
Moving on to the second column, as the complexity of objects in the image increases, the negative effect of overfitting shows and tends to place elements at the upper center, just like most training data.
When the distribution of objects becomes dispersed and not centered, as in the third and fourth columns, or when there is very little available space, as in the fifth column, Scan-and-Print consistently generates visually appealing layouts, making full use of all available element types, even those that constitute a very small fraction of the training data, \textit{e.g.}, embellishment.

More visualized results of (1) constrained generation task, discussed in the next paragraph, and (2) mixed-up samples are presented in the supplementary material.

\paragraph{User-specified constraint.}
Following RALF, we also explore constrained generation and demonstrate the adaptability of Scan-and-Print to the \textit{Category $\rightarrow$ Size + Position} task.
As reported in Tab. \ref{tab:res_constrain}, ours maintains a leading position in most metrics compared to RALF, showcasing its versatility to meet real-world needs.
Moreover, Scan-and-Print achieves an average inference time of 267.8 ms per single inference, which is only 70\% that of RALF (384.5 ms).
This efficiency enables users to quickly explore a variety of high-quality layout options for their materials.

\begin{table}[t]
    \centering
    \resizebox{0.99\linewidth}{!}{%
\begin{tabular}{@{\hspace{1pt}}l@{\hspace{3pt}}|cccc|ccc@{\hspace{1pt}}}
\toprule
Method                & $Ove \downarrow$ & $Ali \downarrow$ & $Und_l \uparrow$ & $Und_s \uparrow$ & $Uti \uparrow$  & $Occ \downarrow$ & $Rea \downarrow$ \\
\midrule
CGL-GAN               & 0.0368           & 0.0046           & 0.8643           & 0.5701           & \textbf{0.2256} & 0.1483           & 0.0173           \\
LayoutDM$^\dag$              & 0.2311          & \textbf{0.0019}  & 0.5875           & 0.1764           & 0.1212          & 0.2319           & 0.0324           \\
AutoReg               & 0.0285           & 0.0030           & 0.7752           & 0.4298           & 0.2029          & 0.1348           & 0.0167           \\
RALF                  & \textbf{0.0095}  & 0.0031           & \textbf{0.9687}  & {\ul 0.8982}     & 0.2137    & {\ul 0.1244}     & {\ul 0.0138}     \\
S-and-P (Ours)  & {\ul 0.0136}     & {\ul 0.0025}     & {\ul 0.9659}     & \textbf{0.9525}  & {\ul 0.2173}    & \textbf{0.1161}  & \textbf{0.0131}  \\
\bottomrule
\end{tabular}
}
\vspace{-0.5em}
\caption{Quantitative results of the \textit{C\! $\rightarrow$\! S\! +\! P} constrained generation task on PKU PosterLayout, \textit{annotated} test split.}
\label{tab:res_constrain}
\vspace{-0.5em}
\end{table}

\begin{table}[t]
    \centering
    \resizebox{0.99\linewidth}{!}{%
\begin{tabular}{@{\hspace{2pt}}ccc|cccc|ccc@{\hspace{1pt}}}
\toprule
V       & S       & P       & $Ove \downarrow$ & $Ali \downarrow$ & $Und_l \uparrow$ & $Und_s \uparrow$ & $Uti \uparrow$  & $Occ \downarrow$ & $Rea \downarrow$ \\
\midrule
        &         &         & \textbf{0.0078}  & 0.0028           & 0.9770           & 0.9015           & 0.1834          & 0.1331           & 0.0231           \\
$\surd$ &         &         & 0.0153           & 0.0025           & 0.9781           & 0.9526           & {\ul 0.2091}    & {\ul 0.1274}     & {\ul 0.0192}     \\
        & $\surd$ &         & 0.0110           & 0.0027           & 0.9737           & 0.8999           & 0.1758          & 0.1394           & 0.0208           \\
$\surd$ & $\surd$ &         & 0.0147           & \textbf{0.0020}  & \textbf{0.9855}  & {\ul 0.9696}     & \textbf{0.2102} & 0.1338           & 0.0223           \\
\midrule
$\surd$ & $\surd$ & $\surd$ & {\ul 0.0090}     & {\ul 0.0024}     & {\ul 0.9831}     & \textbf{0.9709}  & 0.1985          & \textbf{0.1162}  & \textbf{0.0181} \\
\bottomrule
\end{tabular}
}
\vspace{-0.5em}
\caption{Ablation study on each component. (V: Vertex-based layout representation, S: Scan procedure, P: Print procedure.)}
\label{tab:abla_major}
\vspace{-0.8em}
\end{table}

\begin{table}[t]
    \centering
    \resizebox{0.99\linewidth}{!}{%
\begin{tabular}{@{\hspace{1pt}}c@{\hspace{3pt}}|c|cccc|ccc@{\hspace{1pt}}}
\toprule
$k$ & \#FLOPs & $Ove \downarrow$ & $Ali \downarrow$ & $Und_l \uparrow$ & $Und_s \uparrow$ & $Uti \uparrow$  & $Occ \downarrow$ & $Rea \downarrow$ \\
\midrule
\textcolor{gray!60}{Full} & \textcolor{gray!60}{2.90G} & \textcolor{gray!60}{0.0101}           & \textcolor{gray!60}{0.0027}           & \textcolor{gray!60}{0.9846}           & \textcolor{gray!60}{0.9770}           & \textcolor{gray!60}{0.1942}          & \textcolor{gray!60}{0.1158}           & \textcolor{gray!60}{0.0182}           \\
\midrule
96  & 1.46G   & \textbf{0.0090}  & {\ul 0.0024}     & {\ul 0.9831}     & {\ul 0.9709}     & {\ul 0.1985}    & \textbf{0.1162}  & {\ul 0.0181}     \\
48  & 0.76G   & 0.0144           & 0.0032           & 0.9771           & 0.9506           & \textbf{0.2090} & {\ul 0.1224}     & \textbf{0.0180}  \\
24  & 0.41G   & {\ul 0.0093}     & \textbf{0.0022}  & \textbf{0.9848}  & \textbf{0.9755}  & 0.1962          & 0.1282           & 0.0192           \\
\bottomrule
\end{tabular}
}
\vspace{-0.5em}
\caption{Ablation study on the scan size $k$. (Full size: 196)}
\label{tab:abla_size_k}
\vspace{-0.5em}
\end{table}

\begin{table}[t]
    \centering
    \resizebox{0.99\linewidth}{!}{%
\begin{tabular}{@{\hspace{1pt}}l|cccc|ccc@{\hspace{1pt}}}
\toprule
Selection & $Ove \downarrow$ & $Ali \downarrow$ & $Und_l \uparrow$ & $Und_s \uparrow$ & $Uti \uparrow$  & $Occ \downarrow$ & $Rea \downarrow$ \\
\midrule
PCC       & \textbf{0.0090}  & \textbf{0.0024}  & 0.9831           & 0.9709           & \textbf{0.1985} & {\ul 0.1162}     & {\ul 0.0180{\small 6}}    \\
COSIM     & {\ul 0.0094}     & {\ul 0.0025}     & {\ul 0.9836}     & {\ul 0.9754}     & {\ul 0.1940}    & 0.1172           & 0.0180{\small 7}          \\
Random    & 0.0131           & 0.0030           & \textbf{0.9897}  & \textbf{0.9828}  & 0.1894          & \textbf{0.1150}  & \textbf{0.0179}  \\
\bottomrule
\end{tabular}
}
\vspace{-0.5em}
\caption{Ablation study on the mixed-up pairs selection strategies. (PCC: Pearson correlation coefficient, COSIM: Cosine similarity)}
\label{tab:abla_sim_strategy}
\vspace{-0.5em}
\end{table}

\begin{table}[t]
    \centering
    \resizebox{0.99\linewidth}{!}{%
\begin{tabular}{@{\hspace{1pt}}c@{\hspace{3pt}}|c|cccc|ccc@{\hspace{1pt}}}
\toprule
$\alpha$ & Rate  & $Ove \downarrow$ & $Ali \downarrow$ & $Und_l \uparrow$ & $Und_s \uparrow$ & $Uti \uparrow$  & $Occ \downarrow$ & $Rea \downarrow$ \\
\midrule
32       & 25\%  & 0.0211           & {\ul 0.0020}     & 0.9804           & 0.9480           & \textbf{0.2256} & \textbf{0.1134}  & {\ul 0.0182}     \\
64       & 50\%  & 0.0113           & \textbf{0.0017}  & 0.9667           & 0.9503           & {\ul 0.2096}    & 0.1187           & 0.0192           \\
128      & 100\% & \textbf{0.0086}  & 0.0026           & \textbf{0.9869}  & \textbf{0.9781}  & 0.1906          & 0.1195           & 0.0184           \\
256      & 200\% & {\ul 0.0090}     & 0.0024           & {\ul 0.9831}     & {\ul 0.9709}     & 0.1985          & {\ul 0.1162}     & \textbf{0.0181}  \\
\bottomrule
\end{tabular}
}
\vspace{-0.5em}
\caption{Ablation study on the augmented sample size $\alpha$ per mini-batch of size 128.}
\label{tab:abla_size_aug}
\vspace{-1.3em}
\end{table}

\subsection{Ablation Study}
To gain insight into our implementation choices, including (1) each component, (2) the scan size $k$, (3) mixed-up pairs selection strategy, and (4) the augmented sample size $\alpha$ per mini-batch, we conduct extensive ablation studies with PKU PosterLayout, \textit{unannotated} test split.

\paragraph{Effectiveness of each component.}
As the comprehensive analysis provided in Tab. \ref{tab:abla_major}, all components of the Scan-and-Print have contributed positively to its performance.
Starting from the first row, which is our baseline with a compact architecture, it has already outperformed the current SOTA approach, \textit{e.g.}, RALF, in almost all metrics.
More concretely, it falls behind only in $Occ\!\downarrow$ by a negligible 0.0003.
This finding remarkably confirms the point we made in Introduction that the model complexity of existing methods has exceeded the size of available data supports, leading to counterproductive outcomes.
More findings are as follows:

\textbf{Row 1 $\rightarrow$ 2.}
When replacing conventional representation with the proposed VLR, an overall improvement is witnessed, especially in $Und_s\!\uparrow$ by 5.7\%.
While this is as expected, since VLR captures the fine-grained structure of layout elements, what impresses us is the coherent improvements in all content metrics, proving the value of the new representation.

\textbf{Row 1 $\rightarrow$ 3.}
When joining the scan procedure, only a slight performance degradation comes with substantial computational cost savings.
This strongly supports our view that the complexity of image perception in the target task is limited, verifying our decision to reduce its cost in pursuit of inference speed.

\textbf{Row 2, 3 $\rightarrow$ 4.}
When involving both VLR and the scan procedure, their advantages are well combined, showing a good trade-off between effectiveness and computational efficiency.

\textbf{Row 4 $\rightarrow$ 5.}
Finally, introducing the print procedure obtains the best results.
It compensates for the slight visual information loss caused by the patch selection and further improves most graphic metrics through the diverse augmented samples.
The only noticeable decrease in $Uti\!\uparrow$ is attributed to the design of image mixing (Sec.\! \ref{subsec:ipm}) that tends to synthesize difficult cases of very few applicable patches, hence the $Uti\!\uparrow$ in augmented samples is lower than the sources.
Notably, the print procedure incurs only a time cost of 3.2 s per mini-batch and 54.8 min over the entire training period.

\paragraph{Exploration of different scan size $k$.}
Tab. \ref{tab:abla_size_k} observes the impact of varying the perceived patch number on the performance.
We experiment with candidates $k\! =\! \{96,48,24\}$ and also the full-size scenario.
Surprisingly, compared to the full size, $k\! =\! 96$ leads to an overall improvement in content metrics, which suggests that perceiving less applicable patches can introduce undesired noises.
In contrast, if $k$ becomes smaller, where insufficient informative patches are selected, a downward trend of content metrics appears as expected.
Another insight is that when the model pays less attention to visual information, layout features become more dominant and improve graphic metrics.
Furthermore, it is important to highlight that all these results consistently outperform RALF.
When $k\! =\! 24$, the cost saving significantly reaches 95.2\% compared to RALF, which requires 8.51G FLOPs.

\paragraph{Different ways of selecting mixed-up pairs.}
Selecting the mixed-up sources can be crucial \cite{kim-2021-ICLR-comixup} to affect the quality and diversity of the augmented data, which in turn the model's performance.
Therefore, we experiment with three different strategies, including two based on patch indices $(\{\mathcal{P}_i\}, \{\mathcal{P}_j\})$, Pearson correlation coefficient (PCC) and cosine similarity (COSIM), as well as a data-agnostic random strategy.
As reported in Tab. \ref{tab:abla_sim_strategy}, PCC achieves the best overall results, particularly excelling in $Uti\!\uparrow$, which is most impacted by the synthesized challenging cases, as analyzed in previous studies.
Since PCC and COSIM tend to select pairs with similar patch indices, their leading position is expected.
Nevertheless, the random strategy achieves amazingly good results in other metrics.
This finding demonstrates the robustness of the proposed mixup operations, which regains the advantages of their original concepts \cite{zhang-2018-ICLR-mixup}, namely, high adaptability and no stringent prerequisites for the existing training data.

\paragraph{Impacts of augmented sample size $\alpha$.}
Last but not least, Tab. \ref{tab:abla_size_aug} provides insights into the augmented sample size per mini-batch.
We experiment with candidates $\alpha\! =\! \{32, 64, 128,$ $256\}$ given a fixed mini-batch size 128.
As observed, the best outcomes are found at $\alpha\! =\! 128$, 
whereas additional improvement is witnessed in all content metrics when $\alpha$ keeps increasing, demonstrating the potential of involving more augmented data for further advancement.
It is worth reiterating that the cost of creating these samples is extremely low.
Even at $\alpha\! =\! 256$, it takes only 3.2 s per mini-batch, as reported in the previous study.
However, the additional samples do consume more GPU memory and result in longer training times, which is a trade-off to consider.
By investing more efforts in the training stage, the proposed Scan-and-Print has successfully demonstrated excellent real-time performance and generalization capabilities in the inference stage.

%% file: sec/5_conclusion.tex
\vspace{-0.3em}
\section{Discussion}
\paragraph{Conclusion.}
This work discussed the common pitfalls of current content-aware layout generation methods, \textit{i.e.}, high computational costs and low generalization ability.
These defects are attributed to the large number of parameters relative to the limited available training data, and the image encoder is further identified as the culprit.
To address these challenges, we presented a compact autoregressive model accompanying the proposed patch-level data summarization and augmentation approach, Scan-and-Print.
Through extensive experiments, we demonstrated that it has achieved new SOTA results across various benchmarks.
Moreover, compared to the previous SOTA method, it has saved 95\% computational cost in image perception, required only 61\% of the parameters, and reduced inference time to 70\% in generation.

\paragraph{Future work.}
We outline two promising directions that continuously bring valuable contributions to the field.
The first one lies in introducing multi-modal content awareness into the scan procedure.
While most existing work, including ours, considers mainly the background images, subsequent research can incorporate semantics of texts to be put in.
This helps directly target the applicable areas adjacent to the relevant objects, enabling layout generation for the visual design of instructional materials.
The second one is to explore different mixup operations in the print procedure.
While mixup has been applied across various domains \cite{bochkovskiy-2020-arXiv-yolov4,yoon-2021-ACL-ssmix}, we introduce it to layout generation for the first time.
We believe the proposed vertex-level mixup can also boost content-agnostic tasks, which is marvelous because it often works the other way around.
Moreover, we hope this work encourages more researchers to explore various operations, such as the simpler box-level mixup, and we look forward to witnessing the benefits they bring.

%% file: sec/X_appendix.tex
\appendix \section{Definitions of Functions in Algorithm 1}
\label{app:function}

The \textsc{\textbf{group-element-id}} and \textsc{\textbf{arrange-severtex-id}} functions invoked in Algorithm \ref{alg:vlr} are defined as follows:

\vspace{-0.5em}
\begin{algorithm}[th]
    \centering
    \vspace{1pt}
    \textbf{Function} \textsc{group-element-id}$(c_{und}, \text{categories}\ c, \text{boxes}\ b)$ \\
\vspace{2pt}
\hrule
    \begin{algorithmic}[1]
        \IF{$c_{und} \notin c$}
        \STATE $G \gets \{i\}_i^n$
        \ELSE
        \STATE $G \gets \{\text{Tree}(i)\}_i^n$ \ \ $\triangleright$ $i$ as the values of root nodes 
        \STATE $U \gets \textsc{asc-sort}(\{i \mid c_i = c_{und}\}_i^n, \textsc{box-area}(b_i))$
        \STATE $N \gets \{i \mid c_i \neq c_{und}\}_i^n$
        \STATE \textit{Compute} $\text{IoU}_{U \cdot U}, \text{IoU}_{U \cdot N}$ \textit{and zero the diagonal}
        \FOR{$(i, j) \in \{(i, j) \mid (i\! >\! j) \wedge (\text{IoU}_{U \cdot U}[i][j]\! >\! \epsilon_U) \}$}
        \STATE $\triangleright$ Underlay $b_j$ is enclosed by underlay $b_i$
        \IF{$G[j]$ is not visited}
        \STATE \textit{Append} $G[j]$ \textit{to} $G[i]$ \textit{and mark} $G[j]$ \textit{visited}
        \ENDIF
        \ENDFOR
        \FOR{$(i, j) \in \{(i, j) \mid (\text{IoU}_{U \cdot N}[i][j]\! >\! \epsilon_N) \}$}
        \STATE $\triangleright$ Non-underlay $b_j$ is enclosed by underlay $b_i$
        \IF{$G[j]$ is not visited}
        \STATE \textit{Initialize node} $j$ \textit{to} $G[i]$ \textit{and mark} $G[j]$ \textit{visited}
        \ENDIF
        \ENDFOR
        \STATE $G \gets \{G[i] \mid G[i] \ \text{is not visited} \}$
        \ENDIF
        \STATE \textbf{return} $G$
    \end{algorithmic}
\end{algorithm}
\vspace{-1.5em}
\begin{algorithm}[th]
    \vspace{1pt}
    \textbf{Function} \textsc{arrange-severtex-id}$(G, \text{sorting weight}\ W)$ 
\vspace{1pt}
\hrule
    \begin{algorithmic}[1]
        \STATE $A \gets \text{Empty List}, D \gets \text{Empty Dictionary}$
        \FOR{i = 1 \TO \textsc{length(G)}}
        \STATE $j \gets 2G[i].\text{value}$
        \STATE \textit{Append} $j$ \textit{to} $A$
        \IF{$G[i]$ is a leaf node}
        \STATE \textit{Append} $j + 1$ \textit{to} $A$
        \ELSE
        \STATE $D[j] = \textsc{arrange-severtex-id}(G[i].\text{childrean}, W)$ \\
        \STATE \textit{Append} $j + 1$ \textit{to} $D[j]$
        \ENDIF
        \ENDFOR
        \STATE $A \gets \textsc{asc-sort}(A, W[A[i]])$
        \STATE \textit{Extend every} $D[\text{k}]$ \textit{into} $A$ \textit{at} $(\textsc{index-of}(\text{k})+1)\textsc{in}(A)$ 
        \STATE \textbf{return} $A$
    \end{algorithmic}
\end{algorithm}

%% file: supp_sec/0_intro.tex
\section{Supplementary Materials}

This supplementary material provides additional information about the proposed approach, Scan-and-Print.
Sec. \ref{sec:suppl_const} presents more results of the constrained generation task.
Sec. \ref{sec:supp_mixup} presents more examples of mixed-up samples $(\Tilde{I}, \Tilde{L})$ synthesized by the print procedure.

\renewcommand\thetable{\Alph{table}}
\renewcommand\thefigure{\Alph{figure}}
\setcounter{table}{0}
\setcounter{figure}{0}

\begin{figure*}[t!]
    \centering
    \includegraphics[width=\linewidth]{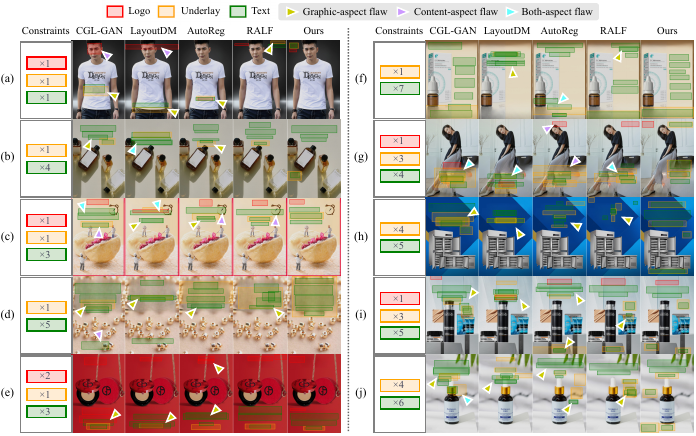}
    \caption{Comparisons of visualized results of the \textit{C\! $\rightarrow$\! S\! +\! P} constrained generation task on PKU PosterLayout, \textit{annotated} test split.}
    \label{fig:vis_c}
\end{figure*}

%% file: supp_sec/1_results.tex
\subsection{Results of Constrained Generation Task}
\label{sec:suppl_const}

Fig. \ref{fig:vis_c} shows the layouts generated by different methods under given categories.
These results illustrate that Scan-and-Print can generate layouts complying with user-specific constraints for various poster background images.
Regardless of the complexity of constraints, it consistently organizes elements into high-quality results.
From the first column, (a), to the last one, (j), the number of given constraints monotonically increases, which makes the task progressively more challenging and simultaneously highlights our approach's advantage.
Specifically, it excels at arranging underlays and their enclosed elements, as seen in (d), where it perfectly places five texts within an underlay without any unpleasant overlays.
Moreover, while some methods fail in difficult cases, such as CGL-GAN, LayoutDM, and AutoReg in (g) as well as RALF in (h), ours still obtains coherent and appropriate structure.
Overall, we demonstrate that Scan-and-Print is highly robust and effective in diverse scenarios, meeting real-world requirements.

%% file: supp_sec/2_samples.tex
\subsection{Examples of Mixed-up Samples}
\label{sec:supp_mixup}

\begin{figure*}[t!]
    \centering
    \includegraphics[width=0.875\linewidth]{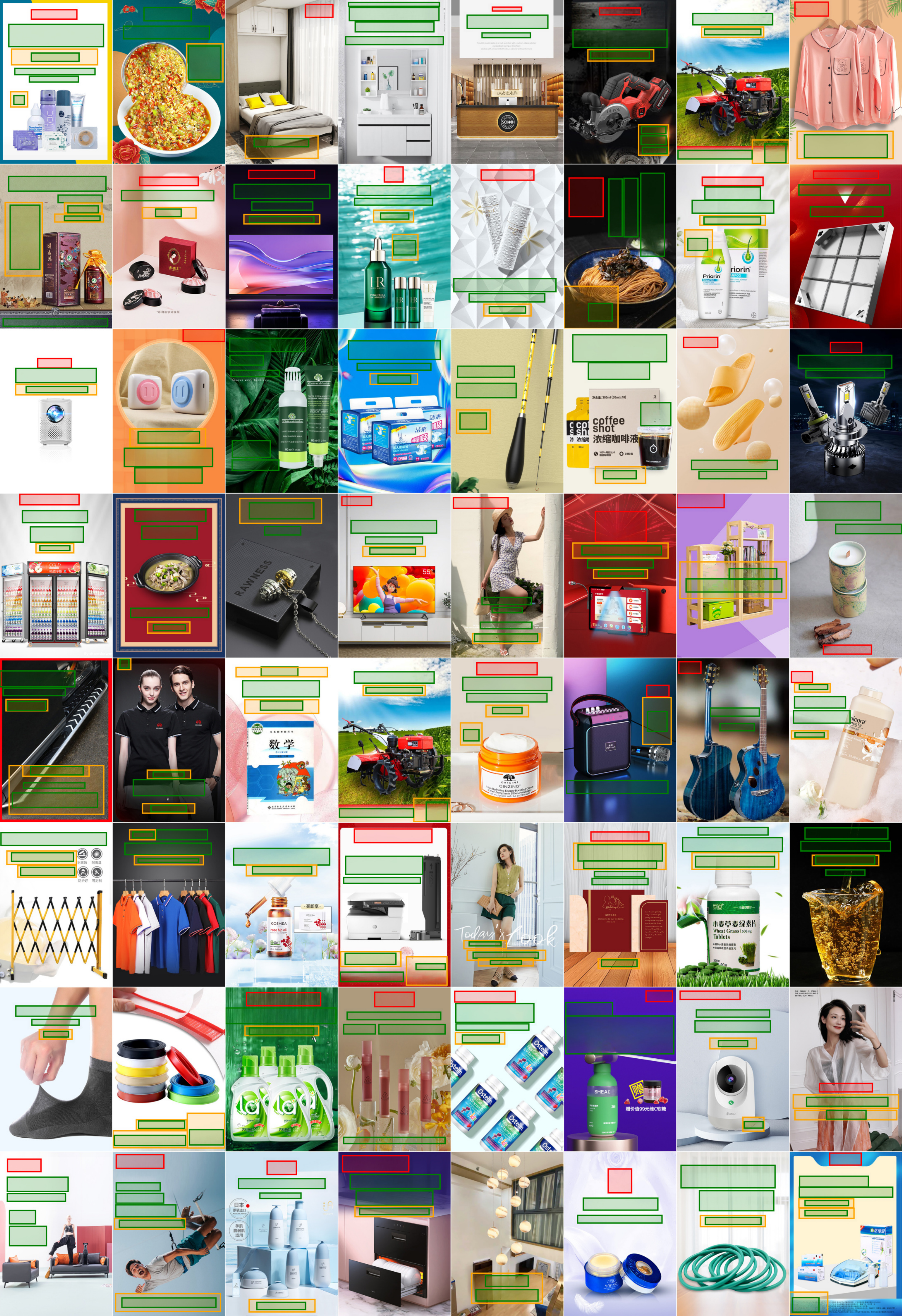}
    \vspace{-0.5em}
    \caption{Examples of the input mini-batch.}
    \label{fig:batch}
    \vspace{-1.2em}
\end{figure*}

\begin{figure*}[t!]
    \centering
    \includegraphics[width=0.875\linewidth]{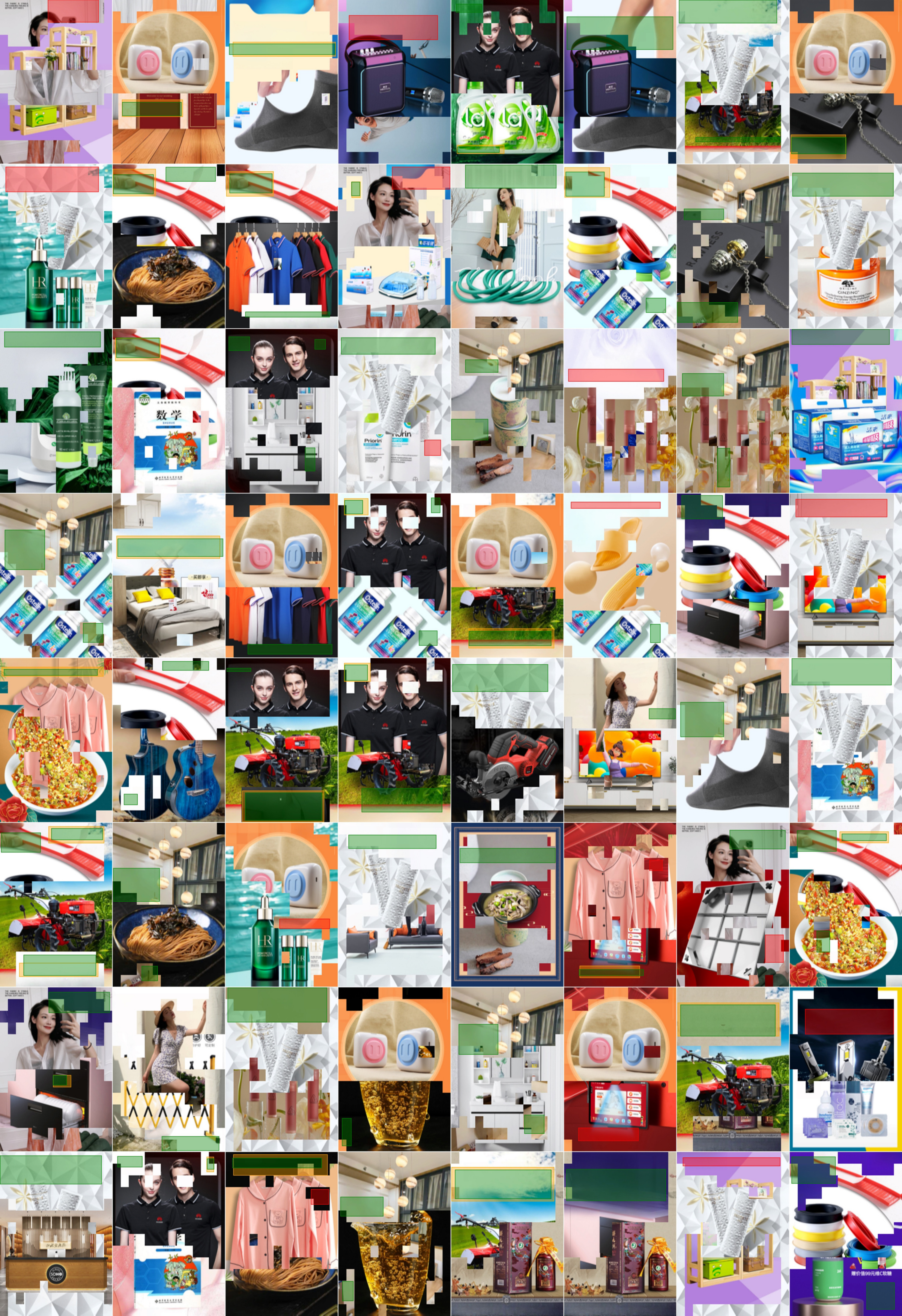}
    \vspace{-0.5em}
    \caption{Examples of the mixed-up samples synthesized by the proposed print procedure.}
    \label{fig:samples}
    \vspace{-1.2em}
\end{figure*}

Given a mini-batch with 64 image-layout pairs, as shown in Fig. \ref{fig:batch}, the print procedure synthesizes various challenging mixed-up samples, a part of which is visualized in Fig. \ref{fig:samples}.

